\documentclass{article}
\usepackage{algorithm}
\usepackage{algpseudocode}
\usepackage{amsmath}
\usepackage{booktabs}
\usepackage{graphicx}
\usepackage{hyperref}
\usepackage{multirow}
\usepackage{xcolor}
\usepackage{spconf}
\usepackage{caption}
\newcommand{\sd}[1]{$_{\pm #1}$}
\title{RoleHear: Pseudo-Labeling for Semi-Supervised Role-Attributed Streaming Speech Recognition}
\name{Zefang Liu, Chenyang Zhu, Sangwoo Cho, Xujun Peng, Rongqing Huang, Shi-Xiong Zhang, Sambit Sahu}
\address{Capital One, USA\\zefang.liu@capitalone.com}
\begin{document}
\ninept
\maketitle
\begin{abstract}
Streaming automatic speech recognition transcribes at low latency but produces an unattributed word stream, while real-time applications such as customer-service analytics need each word tied to a speaker role. Role attribution is usually a separate stage, either an offline or streaming speaker-diarization model aligned to the words or a text-based large language model applied to the transcript, and role labels for training are scarce. We present RoleHear, a semi-supervised pipeline that makes a single cache-aware streaming transducer emit roles inline through lightweight in-vocabulary role tokens: a fast turn-boundary marker followed by a per-role token a few words later. An offline transducer teacher, fine-tuned into a role-attributed recognizer, relabels unlabeled audio, and a multimodal large language model supplies silver role labels where gold annotations are unavailable. The streaming student is fine-tuned on the mixture in a single pass. Across customer-service calls with gold roles and earnings calls with silver roles, RoleHear attributes roles accurately while preserving transcription quality and narrowing the gap to the offline teacher.
\end{abstract}
\section{Introduction}

Pretrained streaming automatic speech recognition (ASR) models, such as the cache-aware FastConformer Recurrent Neural Network Transducer (RNN-T)~\cite{rekesh2023fast,noroozi2024stateful}, deliver low-latency transcription for real-time applications. Their output, however, is an unattributed stream of words. Real-time call-center applications such as live summarization~\cite{chawla2026lessons} and agent assistance need each word tied to a conversational role (for example, customer versus agent) inline with the transcript, not as an after-call step.

Role attribution has traditionally been a separate stage on top of the transcript: an offline or streaming speaker-diarization model aligned to the words, or a post-hoc text-based large language model (LLM). Such approaches build on speaker diarization and speaker-attributed ASR (SA-ASR)~\cite{elshafey2019joint,kanda2020joint,park2024sortformer,shi2024advancing}, which target speaker identity under overlapping speech with auxiliary machinery (a diarization model, a speaker inventory, or serialized decoding) hard to reconcile with cache-aware streaming. Emitting roles inline from a single streaming recognizer avoids this multi-stage complexity but is largely unexplored. Moreover, role annotations are costly to obtain at scale, whereas in-domain unlabeled audio is abundant.

In this work, we present \textbf{RoleHear}, a semi-supervised pipeline for role-attributed streaming ASR, with three contributions. First, we introduce \textbf{decoupled role emission} on a cache-aware streaming transducer: a fast turn-boundary marker followed by a per-role token emitted a few words later, separating turn detection from role commitment so a single streaming model emits roles inline without a separate diarizer. Second, we obtain role-attributed targets by \textbf{single-pass pseudo-labeling}: an offline transducer teacher, fine-tuned into a role-attributed recognizer, relabels unlabeled audio, and the streaming student is fine-tuned on the resulting mixture of labeled and pseudo-labeled data without iteration. Third, we generate \textbf{silver role labels} for transcript-only corpora with a multimodal large language model (MLLM) that conditions on chunk audio and the surrounding transcript context, validated against gold labels, making public data usable without human role annotation. We evaluate RoleHear on customer-service calls with gold roles and earnings calls with silver roles, measuring transcription quality, role accuracy, and the role-latency trade-off.

\section{Related Work}

Attributing words to speakers, the ``who said what'' problem, has been studied as SA-ASR and joint diarization. El Shafey et al.~\cite{elshafey2019joint} add speaker-role tokens to an RNN-T output vocabulary, the closest precedent for RoleHear's in-vocabulary role tokens. Serialized Output Training (SOT) underlies a line of end-to-end SA-ASR for overlapped speech~\cite{kanda2020joint}, later extended to streaming with token-level speaker embeddings~\cite{kanda2022streaming}. A complementary direction couples ASR with dedicated diarizers such as Sortformer and its streaming variant~\cite{park2024sortformer,medennikov2025streaming}, or applies LLMs to speaker attribution~\cite{chen2025who,aronowitz2026speaker}. These methods target speaker identity under overlap and rely on speaker inventories, serialized decoding, or a separate diarizer. For supervision, RoleHear draws on pseudo-labeling, a dominant paradigm for semi-supervised ASR~\cite{kahn2020self,chen2020semi} formalized by iterative pipelines~\cite{xu2020iterative,likhomanenko2021slimipl} and refined with moving-average or momentum teachers~\cite{manohar2021kaizen,higuchi2021momentum} and dynamic caches~\cite{tadevosyan2025unified}. Pseudo-labeling has also been applied to neural diarization~\cite{takashima2021semi}. The closest streaming precedents pair a large offline teacher with a small streaming student through noisy-student training~\cite{hwang2022pseudo} or distill Whisper into a streaming transducer trained from scratch~\cite{thorbecke2024fast}, but neither targets cache-aware streaming~\cite{noroozi2024stateful} nor speaker roles. ReHear~\cite{liu2026rehear} and StreamHear~\cite{liu2026streamhear} apply audio-aware LLM correction and single-pass domain-adapted pseudo-labeling only to plain transcription. Departing from these paradigms, RoleHear attributes a closed set of roles with in-vocabulary role tokens on a cache-aware streaming transducer, adapted by single-pass pseudo-labeling and requiring no speaker enrollment, clustering, diarizer, or iterative refinement.

\section{Methodology}

RoleHear adapts a pretrained cache-aware streaming transducer into a role-attributed ASR (RA-ASR) model through single-pass pseudo-labeling. We denote the offline teacher as $M_T$, the streaming student as $M_S$, and the MLLM used for silver labeling as $M_L$. The data comprise a gold-role set $D_G = \{(x, y, r)\}$ of audio, transcript, and roles, a transcript-only set $D_O = \{(x, y)\}$ whose roles are missing, and an unlabeled pool $D_U = \{x\}$. As summarized in Algorithm~\ref{alg:rolehear}, the pipeline labels $D_O$ with $M_L$, fine-tunes the teacher into an RA-ASR model, generates role pseudo-labels on $D_U$, encodes the role tokens at a fixed word-delay, and fine-tunes the streaming student on the mixture, running each step once.

\begin{algorithm}[h]
\caption{RoleHear: Pseudo-Labeling for Streaming RA-ASR}
\label{alg:rolehear}
\begin{algorithmic}[1]

\Statex \textbf{Models:} Offline teacher $M_T$, streaming student $M_S$, MLLM $M_L$
\Statex \textbf{Data:} Gold-role $D_G = \{(x, y, r)\}$, transcript-only $D_O = \{(x, y)\}$, unlabeled $D_U = \{x\}$

\vspace{0.1cm}
\State \textcolor{gray}{\textit{// Silver role labeling}}
\State $D_S \leftarrow \{ (x, y, M_L(x, y, c)) \mid (x, y) \in D_O \}$

\State \textcolor{gray}{\textit{// Teacher role encoding and fine-tuning}}
\State Encode role tokens in $D_G \cup D_S$, then fine-tune $M_T$

\State \textcolor{gray}{\textit{// Role pseudo-label generation}}
\State $D'_U \leftarrow \{ (x, M_T(x)) \mid x \in D_U \}$

\State \textcolor{gray}{\textit{// Student role-token encoding at word-delay $n$}}
\State Encode $\langle\text{role}\rangle$ and delayed per-role tokens in $D_G \cup D_S \cup D'_U$

\State \textcolor{gray}{\textit{// Student fine-tuning}}
\State Fine-tune $M_S$ on the encoded mixture

\State \textbf{return} $M_S$

\end{algorithmic}
\end{algorithm}

\textbf{Decoupled role emission.} We express streaming role attribution as two in-vocabulary tokens~\cite{elshafey2019joint} on the cache-aware transducer~\cite{noroozi2024stateful,rekesh2023fast}. At each turn boundary the model emits a fast $\langle\text{role}\rangle$ marker, and $n$ words later a per-role value token ($\langle\text{role:customer}\rangle$, $\langle\text{role:agent}\rangle$, and so on). The marker fires on the acoustic turn change and requires no lexical lookahead, whereas the value token is deferred so the role decision can draw on $n$ words of in-turn context, and decoupling the two lets turn detection proceed without waiting on role commitment. Targets are built deterministically, inserting $\langle\text{role}\rangle$ at each boundary and the per-role token $n$ words later, a fixed word-delay $n$ that leaves the word stream unchanged. For example, at delay $n = 2$ an agent turn and the customer's reply are encoded as
\[
\begin{aligned}
&\langle\text{role}\rangle\;\;\text{Thank you}\;\;\langle\text{role:agent}\rangle\;\;\text{for calling.}\\
&\langle\text{role}\rangle\;\;\text{I have}\;\;\langle\text{role:customer}\rangle\;\;\text{a question.}
\end{aligned}
\]
with each $\langle\text{role}\rangle$ marker at a turn boundary and its per-role token two words later, while $n = 0$ puts the role token right after the marker, and $n < 0$ places it relative to the turn's end. A non-negative delay keeps the student's targets causal, while the full-context teacher may place the token turn-final ($n=-1$).

\textbf{Silver role labeling.} For corpora with gold transcripts but no role labels, we recover roles without human annotation using the MLLM $M_L$ (Figure~\ref{fig:role_prompt}). For each chunk, $M_L$ is prompted with the chunk audio, the chunk transcript as a numbered token list, and the transcripts of the neighboring chunks on both sides, and returns a list of role segments, each a role from the closed set and the token index at which it starts; slicing the transcript at these boundaries gives contiguous role-labeled spans, so a chunk may carry more than one role. Each role is decided from both the chunk audio (voice and tone) and the words, including how each speaker refers to themselves and others, with the neighboring transcripts keeping assignments consistent across chunk boundaries. Because only the target chunk carries audio, the prompt marks that chunk in the context and states that its neighbors are text alone. Where the recording is stereo, each context line is tagged with its channel and the roles are restricted to those of the target channel; the prompt drops this rule on mono recordings. Conditioning on the chunk audio, not the transcript alone, separates this from text-only LLM role assignment~\cite{chen2025who}. The procedure needs only a transcript, so it labels transcript-only corpora to form the silver-role set $D_S$ and, applied to the gold-labeled set, is scored against the gold roles to estimate its reliability.

\begin{figure}[t]
\centering
\footnotesize
\fbox{\parbox{0.92\columnwidth}{
\textbf{System:} You label speaker roles in a recording. You are given the audio of one chunk, that chunk's transcript as a numbered list of tokens (\texttt{[index] token}), and the neighboring chunks' transcripts for context. Split the chunk into consecutive segments, each spoken by a single role, and report where each segment starts.\\[3pt]
\textit{Rules:} (1)~The audio is the chunk to label (marked \texttt{>>>} in the context), not the neighboring chunks, which are given as text only. (2)~Decide each segment's role from both the audio (voice and tone) and what is said: the words themselves and how each speaker refers to themselves and to others; use the neighboring transcripts to stay consistent across the recording. (3)~When the chunk is marked with a channel, use only roles that belong to that channel. (4)~Report one entry per segment: its role and the token index where it starts; the first segment starts at index 0, and a new segment begins only where the role changes, otherwise keep the chunk as one segment. (5)~Do not change the transcript; report only indices and roles, and use only the roles listed below (never invent one).\\[3pt]
\textit{Allowed roles:} One \texttt{role: description} line per role of the closed set.\\[3pt]
\textbf{User:} The neighbor context in time order, one line per chunk as \texttt{Channel <c> [MM:SS - MM:SS]: <text>}, with \texttt{>>>} marking the chunk to label, then that chunk's channel and its numbered tokens \texttt{[0] Thank [1] you [2] for [3] calling. [4] I [5] have [6] a [7] question.} The chunk audio is attached to this turn.\\[3pt]
\textbf{Output:} \texttt{role\_segments}, a list of entries, each giving a \texttt{role} and the \texttt{start} token index, constrained to the role set by a JSON schema. The chunk above returns \texttt{[(agent, 0), (customer, 4)]}.
}}
\caption{Role-labeling prompt for the MLLM $M_L$: a system instruction carrying the closed role set, a user turn with neighbor context and the target chunk's numbered tokens and audio, and a schema-constrained role-segment output.}
\label{fig:role_prompt}
\end{figure}

\textbf{Teacher fine-tuning and pseudo-labels.} The offline teacher $M_T$, a full-context transducer, extends the base vocabulary with the $\langle\text{role}\rangle$ marker and the per-role tokens, adding embeddings for the new tokens without re-initializing the pretrained parameters. Fine-tuning it on the union of gold- and silver-role data $D_G \cup D_S$ under the transducer loss yields an offline RA-ASR model. Because it sees the whole utterance, it commits roles at turn granularity without a latency constraint. It then transcribes the unlabeled pool $D_U$ once with greedy decoding, producing role-attributed pseudo-labels $D'_U$ that extend supervision to the abundant in-domain audio.

\textbf{Student fine-tuning.} The streaming student $M_S$ extends its vocabulary the same way. After encoding role tokens at the target word-delay $n$, it is fine-tuned on the mixture $D_G \cup D_S \cup D'_U$ under the standard RNN-T loss, in a single pass without iteration. The gold and silver labels anchor the role tokens on trusted transcripts, while the teacher's pseudo-labels supply the scale of in-domain audio needed to learn them, so one fine-tuning run adapts transcription and role attribution together. At inference the student decodes in cache-aware streaming mode and emits the $\langle\text{role}\rangle$ marker and per-role tokens inline with the word stream, adding no model beyond the recognizer.

\section{Experiments}

This section describes the datasets and their role labels, the training and inference setup, the main role-attributed ASR comparison across the teacher and streaming student, and two ablations on the role-emission delay and the right context.

\subsection{Datasets}

We evaluate on two public earnings-call corpora with silver role labels and one proprietary call-center corpus with gold role labels; Table~\ref{tab:datasets} lists per-split hours. \textbf{Earnings-21}~\cite{del2021earnings} is 44 quarterly earnings calls across nine financial sectors with entity-dense transcripts, and \textbf{Earnings-22}~\cite{del2022earnings} is 125 English earnings calls from global companies spanning seven world-English regions, of which four predominantly non-English files are dropped at prepare time, leaving 121. Both are transcript-only, so an MLLM silver labeler, conditioning on each chunk's audio and the neighboring-chunk transcripts, assigns their three native roles: operator (who runs the call), presenter (the company speakers), and questioner (the analysts). We reuse the chunk-level preparation of StreamHear~\cite{liu2026streamhear}: voice-activity detection (VAD) segmentation, CTC-Segmentation forced alignment, and prior-regularized dynamic-programming realignment. \textbf{ServiceCall} is a proprietary human-annotated set of stereo customer-service calls (channel~2 the primary agent, channel~1 the customer and any other parties) with gold speaker-role labels and chunk-level timestamps. From the gold roles we form a 3-role set, customer, agent, and system (the automated prompts), and a 5-role set that splits the agent into a primary agent (the lead), an internal agent (other staff), and an external agent (a third party). Its unlabeled pool draws additional in-domain calls segmented by VAD.

\begin{table}[h]
\centering
\caption{Dataset statistics, reporting the total duration in hours per split and the average chunk length in seconds.}
\label{tab:datasets}
\footnotesize
\begin{tabular}{lrrr}
\toprule
\textbf{Split} & \textbf{Earnings-21} & \textbf{Earnings-22} & \textbf{ServiceCall} \\
\midrule
Labeled Train & 10.72 & \phantom{0}24.76 & \phantom{0}31.08 \\
Labeled Val & \phantom{0}3.09 & \phantom{00}9.16 & \phantom{00}7.22 \\
Labeled Test & \phantom{0}4.88 & \phantom{0}12.38 & \phantom{00}6.86 \\
Unlabeled & 18.85 & \phantom{0}63.01 & \phantom{0}62.56 \\
\midrule
\textit{Total} & 37.54 & 109.31 & 107.72 \\
\textit{Chunk (s)} & 23.3 & \phantom{0}23.0 & \phantom{00}9.8 \\
\bottomrule
\end{tabular}
\end{table}

\subsection{Experimental Setup}

\textbf{Models.} The offline teacher $M_T$ is Parakeet-TDT-0.6B-v3~\cite{rekesh2023fast}, a 0.6B-parameter full-context FastConformer transducer with token-and-duration decoding. The streaming student $M_S$ is Nemotron-Speech-Streaming-EN-0.6B~\cite{noroozi2024stateful}, a 0.6B cache-aware FastConformer-RNN-T trained with multi-latency right-context (RC) sampling; unless noted we decode at $\text{RC}{=}1$ (an $80$ ms chunk plus one $80$ ms lookahead frame, for $160$ ms of algorithmic latency). The silver labeler $M_L$ is Gemma-4-12B-IT~\cite{team2026gemma}, a multimodal LLM prompted per chunk with the chunk audio and neighboring-chunk text within $5$ minutes on each side. Both recognizers extend their vocabularies with the $\langle\text{role}\rangle$ marker and the per-role tokens of Algorithm~\ref{alg:rolehear}, grafting fresh rows for the new tokens while preserving the pretrained parameters; the new tokens' joint-output bias is initialized to $-100$ so they stay silent until fine-tuning moves them.

\textbf{Training.} All fine-tuning runs share the same recipe: AdamW optimizer with learning rate $5\times 10^{-4}$, betas $[0.9, 0.98]$, weight decay $10^{-3}$; cosine schedule with $10\%$ warmup and minimum learning rate $10^{-6}$; bf16 precision; SpecAugment ($2$ frequency masks $\times$ $10$ time masks). Effective batch size $64$ ($4$ A100-SXM-40GB GPUs, per-GPU batch $4$, gradient accumulation $4$) for the streaming student; teacher fine-tuning uses per-GPU batch $2$ with gradient accumulation $8$ to fit the transducer joint on Earnings-length chunks. Training runs $20$ epochs on ServiceCall and Earnings-22, and $50$ on Earnings-21, whose smaller labeled split yields fewer optimizer steps per epoch. Checkpoints are selected on validation RWER (Role-attributed WER, introduced below), so role attribution drives selection. Each configuration is repeated for $5$ seeds and we report mean and standard deviation. All training and inference use the NVIDIA NeMo framework~\cite{kuchaiev2019nemo}.

\textbf{Inference and metrics.} Inference uses greedy decoding, full context for the teacher and under the cache-aware streaming attention mask for the student. All metrics are computed after a text-normalization pipeline: lowercasing; normalizing numbers, contractions, spelling, and compounds; and removing bracketed annotations, diacritics, filler words, and punctuation. Word error rate (WER) measures transcription quality. Role-attributed WER (RWER) measures role attribution carried in the word stream, following speaker-attributed WER~\cite{kanda2020joint} with roles named rather than clustered: for each utterance $u$ we bucket the reference and hypothesis words by role $r \in \mathcal{R}$ and score each role's reference words against its hypothesis words, giving
\begin{equation}
\text{RWER} = \frac{\sum_u \sum_{r \in \mathcal{R}} (S_{u,r} + I_{u,r} + D_{u,r})}{\sum_u \sum_{r \in \mathcal{R}} N_{u,r}},
\end{equation}
where $S_{u,r}$, $I_{u,r}$, $D_{u,r}$ are the substitutions, insertions, and deletions between role $r$'s reference and hypothesis words and $N_{u,r}$ is its reference word count; a role present on only one side contributes pure insertions or deletions, and, because words are scored per role within each utterance, a cross-turn role swap counts as errors in both roles rather than canceling. We also report macro-averaged precision, recall, and F1 over the decoded per-word role labels, with precision and recall computed per role and the per-role F1 averaged equally across roles so rare roles weigh as much as frequent ones. For the streaming student we additionally report the algorithmic latency (AL), $(1{+}\text{RC})\times 80$ ms.

\subsection{Experimental Results}

\begin{table*}[t]
\centering
\caption{Role-attributed ASR on the labeled test split (\%, $5$-seed mean, std in subscript). ServiceCall-5R uses the $5$-role set, the others the $3$-role set; roles are gold on ServiceCall, silver on Earnings. The student rows decode at $\text{RC}{=}1$ ($160$ ms). FT denotes fine-tuned. Earnings columns are absolute; ServiceCall columns are the change ($\Delta$, in pp) from the Teacher + MLLM row.}
\label{tab:main}
\small
\resizebox{\textwidth}{!}{%
\begin{tabular}{lcccccccccccc}
\toprule
\multirow{2.5}{*}{\textbf{Method}} & \multicolumn{3}{c}{\textbf{Earnings-21}} & \multicolumn{3}{c}{\textbf{Earnings-22}} & \multicolumn{3}{c}{\textbf{ServiceCall-3R}} & \multicolumn{3}{c}{\textbf{ServiceCall-5R}} \\
\cmidrule(lr){2-4} \cmidrule(lr){5-7} \cmidrule(lr){8-10} \cmidrule(lr){11-13}
& \textbf{WER}\,$\downarrow$ & \textbf{RWER}\,$\downarrow$ & \textbf{F1}\,$\uparrow$ & \textbf{WER}\,$\downarrow$ & \textbf{RWER}\,$\downarrow$ & \textbf{F1}\,$\uparrow$ & \textbf{$\Delta$WER}\,$\downarrow$ & \textbf{$\Delta$RWER}\,$\downarrow$ & \textbf{$\Delta$F1}\,$\uparrow$ & \textbf{$\Delta$WER}\,$\downarrow$ & \textbf{$\Delta$RWER}\,$\downarrow$ & \textbf{$\Delta$F1}\,$\uparrow$ \\
\midrule
Teacher + MLLM & 6.49\sd{0.00} & - & - & 12.24\sd{0.00} & - & - & \phantom{-}0.00\sd{0.00} & \phantom{-}0.00\sd{0.00} & \phantom{-}0.00\sd{0.00} & \phantom{-}0.00\sd{0.00} & \phantom{-}0.00\sd{0.00} & \phantom{-}0.00\sd{0.00} \\
FT Teacher & 6.88\sd{0.06} & 19.24\sd{0.60} & 82.83\sd{0.70} & 11.79\sd{0.04} & 25.94\sd{0.78} & 80.59\sd{0.96} & -2.98\sd{0.06} & -6.29\sd{0.44} & +1.21\sd{0.77} & -3.04\sd{0.07} & -5.56\sd{0.39} & -8.92\sd{2.64} \\
\midrule
Student + MLLM & 8.27\sd{0.00} & - & - & 13.55\sd{0.00} & - & - & +0.78\sd{0.00} & +1.66\sd{0.00} & -1.20\sd{0.00} & +0.78\sd{0.00} & -1.82\sd{0.00} & +1.42\sd{0.00} \\
FT Student & 7.74\sd{0.14} & 31.78\sd{0.63} & 64.06\sd{1.66} & 12.93\sd{0.06} & 42.29\sd{0.71} & 52.16\sd{1.65} & -1.31\sd{0.17} & +0.21\sd{0.65} & -8.83\sd{1.39} & -1.30\sd{0.09} & +2.81\sd{0.45} & -19.83\sd{0.93} \\
RoleHear & 7.56\sd{0.09} & 27.55\sd{0.81} & 72.95\sd{2.20} & 12.33\sd{0.16} & 35.80\sd{0.60} & 65.97\sd{2.07} & -1.99\sd{0.09} & -2.92\sd{0.35} & -5.95\sd{0.83} & -2.10\sd{0.13} & -1.80\sd{0.98} & -17.34\sd{1.82} \\
\bottomrule
\end{tabular}%
}
\end{table*}

Table~\ref{tab:main} pairs each pretrained recognizer with the MLLM labeler $M_L$ over its ASR hypothesis as a cascade baseline (Teacher + MLLM, Student + MLLM, a second offline pass), alongside the two fine-tuned recognizers and the streaming RoleHear student. We report absolute WER, RWER, and F1 on the public Earnings sets, and, on the proprietary ServiceCall, their percentage-point change ($\Delta$) against the Teacher + MLLM cascade. The silver labels are reliable: $M_L$ scores macro-F1 above $92\%$ and accuracy above $94\%$ on ServiceCall gold at three and five roles, so we apply it to Earnings without gold supervision. Because that reference is $M_L$'s own output, the cascade rows show Earnings WER only, leaving RWER and F1 blank to avoid scoring $M_L$ against itself. Fine-tuning the offline teacher into a role-attributed recognizer barely affects transcription, adding at most $0.4$ percentage points (pp) of WER over the pretrained teacher while attributing roles accurately, with macro F1 above $80\%$ on both Earnings sets. The streaming student fine-tuned with the same tokens (FT Student) keeps WER at or below its pretrained counterpart. RoleHear then improves its role attribution in a single pass, lowering RWER by $4.2$ and $6.5$ pp and raising macro F1 by $8.9$ and $13.8$ pp on Earnings-21 and Earnings-22 at equal or better WER. On ServiceCall, RoleHear beats the Teacher + MLLM cascade on WER and RWER at both role sets in a single streaming pass, trailing only on F1 (the cascade runs the 12B labeler offline), and it improves on the supervised FT Student by $3.1$ and $4.6$ pp RWER and $2.9$ and $2.5$ pp F1.

\subsection{Ablation Studies}

We run two ablations on the streaming student, both on Earnings-21: its role-emission delay and its right context, which set the role latency and algorithmic latency, respectively.

\subsubsection{Role-Emission Delay}

We sweep the role-token delay on Earnings-21 for both the offline teacher and the streaming student, the student decoded at $\text{RC}{=}1$ (Table~\ref{tab:delay}). The offline teacher has no corpus-consistent optimum: Earnings-21 mildly favors delay $0$, while Earnings-22 favors delay $-1$ decisively ($25.9\%$ RWER at delay $-1$ vs $32.3\%$ at delay $0$, matched WER, not shown). Full context lets it label a turn from either end, and as only the Earnings-22 gap is large, we place the teacher token at delay $-1$. The streaming student, which must commit to a role before the turn ends, behaves oppositely: emitting at the boundary (delay $0$) collapses attribution ($46.1\%$ F1) because no turn-internal words have decoded yet, while a small positive delay restores it, delays $1$ and $3$ tying for the best RWER (near $31.8\%$). Delay $-1$ is best on F1 but worst on RWER, and releases the role only after the turn ends, the highest latency of all settings. They diverge because macro-F1 scores only the per-word role prediction and tolerates word errors, whereas RWER is a role-bucketed WER and demands the right words under the right roles. WER stays flat throughout, so we adopt delay $1$ for the student, the smallest value that avoids collapse and thus the lowest role latency (RL) at the best RWER: ${\approx}360$ ms, the mean word duration.

\begin{table}[h]
\centering
\caption{Role-emission delay on Earnings-21, fine-tuning only; student at $\text{RC}{=}1$. WER, RWER, F1 (\%); role latency $\text{RL}\approx\text{delay}\times 360$ ms. Delay $-1$ places the token after the turn's last word (highest latency). \textbf{Bold}/\underline{underline}: best/second-best per column (student rows).}
\label{tab:delay}
\footnotesize
\begin{tabular}{llcccc}
\toprule
\textbf{Model} & \textbf{Delay} & \textbf{RL} & \textbf{WER}\,$\downarrow$ & \textbf{RWER}\,$\downarrow$ & \textbf{F1}\,$\uparrow$ \\
\midrule
\multirow{2}{*}{Teacher} & \phantom{$-$}0 & 0 & 6.88\sd{0.08} & \textbf{17.99}\sd{0.85} & \textbf{83.80}\sd{1.44} \\
 & $-1$ & Turn & \textbf{6.88}\sd{0.06} & 19.24\sd{0.60} & 82.83\sd{0.70} \\
\midrule
\multirow{5}{*}{Student} & \phantom{$-$}0 & 0 & \textbf{7.71}\sd{0.07} & 37.32\sd{1.03} & 46.07\sd{4.90} \\
 & \phantom{$-$}1 & 360 & 7.74\sd{0.14} & \underline{31.78}\sd{0.63} & \underline{64.06}\sd{1.66} \\
 & \phantom{$-$}3 & 1080 & \underline{7.74}\sd{0.12} & \textbf{31.75}\sd{1.51} & 60.91\sd{3.05} \\
 & \phantom{$-$}5 & 1800 & 7.75\sd{0.05} & 32.89\sd{1.22} & 60.10\sd{1.74} \\
 & $-1$ & Turn & 7.81\sd{0.06} & 45.29\sd{3.11} & \textbf{75.91}\sd{1.08} \\
\bottomrule
\end{tabular}
\end{table}

\subsubsection{Right Context}

We then vary the right context at each delay (Table~\ref{tab:context}). Right context sets the algorithmic latency $\text{AL}=(1{+}\text{RC})\times 80$ ms for every word, whereas the role delay adds a role latency $\text{RL}\approx\text{delay}\times 360$ ms for the role token alone. A larger right context helps on every metric at both delays, and at delay $0$ it rescues the boundary-emission collapse, lifting F1 from $42.5\%$ to $65.9\%$ as RC grows from $0$ to $13$. But this gain is paid by the whole stream, since $\text{RC}{=}13$ raises AL from $80$ to $1120$ ms on every word. A role delay is the cheaper source of role quality because it defers only the role token: at $\text{RC}{=}0$, moving from delay $0$ to delay $1$ alone lifts F1 to $63.4\%$ at $80$ ms AL. The two latencies exchange rather than stack, so we keep the student at delay $1$ with a small $\text{RC}{=}1$ ($160$ ms AL).

\begin{table}[h]
\centering
\caption{Right context and delay on Earnings-21, fine-tuning only; streaming student. WER, RWER, F1 (\%); role latency $\text{RL}\approx\text{delay}\times 360$ ms, algorithmic latency $\text{AL}=(1{+}\text{RC})\times 80$ ms. \textbf{Bold}/\underline{underline}: best/second-best per column within each delay.}
\label{tab:context}
\footnotesize
\begin{tabular}{crrrccc}
\toprule
\textbf{Delay} & \textbf{RC} & \textbf{RL} & \textbf{AL} & \textbf{WER}\,$\downarrow$ & \textbf{RWER}\,$\downarrow$ & \textbf{F1}\,$\uparrow$ \\
\midrule
\multirow{4}{*}{0} & 0 & 0 & 80 & 7.84\sd{0.07} & 36.25\sd{1.17} & 42.55\sd{1.42} \\
 & 1 & 0 & 160 & 7.71\sd{0.07} & 37.32\sd{1.03} & 46.07\sd{4.90} \\
 & 6 & 0 & 560 & \underline{7.51}\sd{0.05} & \underline{32.60}\sd{0.87} & \underline{60.98}\sd{8.64} \\
 & 13 & 0 & 1120 & \textbf{7.40}\sd{0.07} & \textbf{29.54}\sd{1.13} & \textbf{65.93}\sd{9.87} \\
\midrule
\multirow{4}{*}{1} & 0 & 360 & 80 & 7.86\sd{0.13} & 32.31\sd{0.51} & 63.39\sd{0.76} \\
 & 1 & 360 & 160 & 7.74\sd{0.14} & 31.78\sd{0.63} & 64.06\sd{1.66} \\
 & 6 & 360 & 560 & \underline{7.49}\sd{0.11} & \underline{31.37}\sd{0.53} & \underline{65.68}\sd{1.37} \\
 & 13 & 360 & 1120 & \textbf{7.36}\sd{0.23} & \textbf{30.31}\sd{0.55} & \textbf{67.18}\sd{1.80} \\
\bottomrule
\end{tabular}
\end{table}

\section{Conclusion}

We presented RoleHear, a semi-supervised pipeline that makes a single cache-aware streaming transducer attribute conversational roles inline, through a fast turn-boundary marker followed by a per-role token a few words later. An offline teacher, fine-tuned into a role-attributed recognizer with the same in-vocabulary tokens, relabels unlabeled audio in one pass, and an MLLM supplies silver role labels where gold annotations are missing, so the streaming student learns transcription and role attribution together, with no separate diarizer or iterative refinement. Across earnings calls with silver roles and customer-service calls with gold roles, adding role tokens is nearly free on transcription, and single-pass pseudo-labeling substantially narrows the streaming student's role-attribution gap to the offline teacher, while a small role-emission delay keeps algorithmic latency low by deferring only the role token, rather than widening the right context and delaying every word.

\bibliographystyle{IEEEbib}
\bibliography{refs}
\end{document}